\documentclass[conference]{ieeeconf}
\IEEEoverridecommandlockouts
\usepackage{cite}
\usepackage{amsmath,amssymb,amsfonts}
\usepackage{algorithmic}
\usepackage{graphicx}
\usepackage{textcomp}
\usepackage{xcolor}
\usepackage{subcaption}
\usepackage{caption}
\usepackage{diagbox}
\usepackage{slashbox}
\usepackage{numprint}

\usepackage{tabularx}
\usepackage[table-number-alignment=center, table-format=1.2]{siunitx}
\usepackage{multicol, multirow}
\usepackage{booktabs}
\usepackage{url}

\def\BibTeX{{\rm B\kern-.05em{\sc i\kern-.025em b}\kern-.08em
    T\kern-.1667em\lower.7ex\hbox{E}\kern-.125emX}}

\DeclareMathOperator*{\argmin}{arg\,min}

\newcommand{\etal}{\textit{et al.}}

\newcolumntype{x}[1]{>{\centering\arraybackslash\hspace{0pt}}p{#1}}

\begin{document}

\title{Accurate Prior-centric Monocular Positioning with Offline LiDAR Fusion \\

\author{Jinhao He$^{1, *}$, Huaiyang Huang$^{2, *}$, Shuyang Zhang$^{3}$, Jianhao Jiao$^{3, 4}$,  Chengju Liu$^{5}$ and Ming Liu$^{1, 6}$}

\thanks{$^{1}$ The authors are with the
		Thrust of Robotics and Autonomous Systems, The Hong Kong University of Science and Technology (Guangzhou),  Nansha District, Guangzhou, China.
        $^{2}$ The author is with the Baidu autonomous driving technology department (ADT).
        $^{3}$ The authors are with the
		Department of Electronic and Computer Engineering, The Hong Kong University of Science and Technology, Clear Water Bay, Kowloon, Hong Kong.
        $^{4}$ The author is with the Department of Computer Science, University College London, Gower Street, WC1E 6BT, London, UK.
        $^{5}$  The author is with the School of Electronics and Information Engineering, Tongji University, Shanghai, China.
        $^{6}$ The author is with the HKUST Shenzhen-Hong Kong Collaborative Innovation Research Institute,  Shenzhen, China.
$^{*}$The first two authors are equally contributed. Corresponding author: Jianhao Jiao {\tt\footnotesize ucacjji@ucl.ac.uk}.}
\thanks{This work was supported by the National Natural Science Foundation of China under Grants (62333017, 62173248), Guangdong Basic and Applied Basic Research Foundation (No. 2021B1515120032), Guangzhou-HKUST(GZ) Joint Funding Program (No. 2024A03J0618), and Project of Hetao Shenzhen-Hong Kong Science and Technology Innovation Cooperation Zone(HZQB-KCZYB-2020083), awarded to Prof. Ming Liu.}

}

\maketitle

\begin{abstract}
Unmanned vehicles usually rely on Global Positioning System (GPS) and  Light Detection and Ranging (LiDAR) sensors to achieve high-precision localization results for navigation purpose. However, this combination with their associated costs and infrastructure demands, poses challenges for widespread adoption in mass-market applications. In this paper, we aim to use only a monocular camera to achieve comparable onboard localization performance by tracking deep-learning visual features on a LiDAR-enhanced visual prior map.
Experiments show that the proposed algorithm can provide centimeter-level global positioning results with scale, which is effortlessly integrated and favorable for low-cost robot system deployment in real-world applications.  

\end{abstract}


\section{Introduction} \label{sec:intro}
A localization module that estimates the robot systems' states, e.g. position and orientation, is fundamental to downstream tasks like navigation, obstacle avoidance, decision-making, etc. From a systematic perspective, the state estimation results need to be represented under a global reference frame to effectively collaborate with perception information or prior annotations like the position of traffic lights and drivable areas. Since the Global Positioning System (GPS) is denied occasionally in places like urban canyons, mainstream solutions usually rely on an accurate global offline map generated before the autonomous operation~\cite{AV_survey}.

In the domain of autonomous driving, usually, a centimeter-level localization precision is preferred. Achieving this level of localization accuracy demands innovative approaches and the fusion of multiple data sources.  LiDAR-based methods~\cite{zhang2014loam, shan2018lego, fastlio2, chen2022dlo} stand out benefiting from the capability of LiDAR to provide high-precision depth measurements. However, LiDAR sensor remains relatively expensive, need higher power consumption, and is more fragile to physical shock when compared to cameras, posing challenges for mass-market deployment.

\begin{figure}[t]
    \centering
    \includegraphics[width=0.438\textwidth]{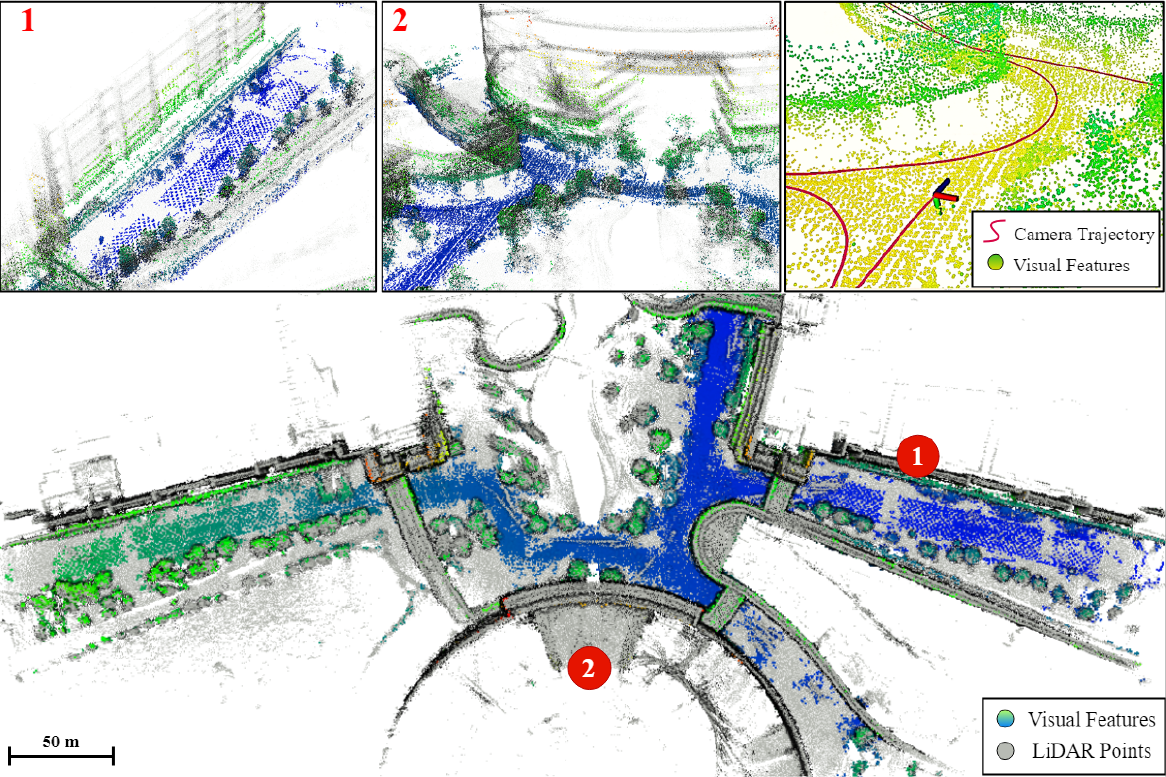}
    \caption{Image tracking in the visual feature map generated by the proposed method (top-right). The feature map can well capture both the visual detail and the precise geometry structure of the environment (top-left, top-mid, bottom).}
    \label{fig:illustrate}
    \vspace{-15pt}
\end{figure}

Cameras on the other hand recover the depth information in an indirect way, which serves as a significant alternative for robot localization. 
Taking visual map construction as the first step, prevalent camera-centric methods extract features such as keypoints, edges, and textures and match them in the image sequence. The camera poses together with the 3D coordinates of the features are estimated by bundle adjustment (BA), this process is also known as structure from motion (SfM)~\cite{ullman1979sfm}. A stereo setup or extra inertial measurement unit (IMU) is needed to help recover the absolute scale of the scene~\cite{qin2018vins, lsdstereo, mur2017orb}. 
The recovered visual map can be used as a reference for further localization to ensure a fixed global frame and reduce drift\cite{oleynikova2015real}.
 Along this track, maplab \cite{schneider2018maplab}  proposed a multi-session visual-inertial mapping framework to improve the efficiency of large-scale visual mapping. Lynen \etal\cite{lynen2015get, lynen2020large} employed the map and descriptor compression scheme to resolve the online computation burden for large-scale, real-time tracking on the visual map.
However, it is still challenging to maintain the computational efficiency and accuracy of both SfM and localizing against a SfM model, especially compared to LiDAR-based methods. For example,  without precise depth measurements, visual localization systems generally exhibit a lower level of precision than LiDAR. However, the rich low-level texture and high-level semantic information captured in the image still make the camera irreplaceable in the robotic system, especially localization.

To narrow the gap between visual-based and LiDAR-based localization methods, a recent branch of research~\cite{caselitz2016monocular,kim2018stereo,yuline,zuo2019visual,huang2020gmmloc,zhangsemantic,cramariuc2022maplab,oishi2020c, oishi2023lc} focuses on leveraging the high-precision LiDAR point cloud in the visual state estimation process. Caselitz \etal~\cite{caselitz2016monocular} tracked the 7 degrees of freedom (DoF) camera pose with scale by reconstructing 3D points from image features and continuously matching them against a given 3D LiDAR map using Iterative Closest Point (ICP) scheme. Kim \etal~\cite{kim2018stereo} used a stereo camera setup and tracked the stereo disparity map against the 3D LiDAR map via minimizing the depth residual. Zuo \etal~\cite{zuo2019visual} proposed to fuse the prior LiDAR map constraints into a  MSCKF stereo-visual inertial framework by performing normal distribution transform (NDT)-based registration between the visual semi-dense reconstruction and prior map. Yu \etal~\cite{yuline} extracted line features from both image and LiDAR map and the camera pose is recovered using 2D-3D line correspondences. Huang \etal~\cite{huang2020gmmloc} modeled the prior map by the Gaussian Mixture Model (GMM). The camera poses and visual structure are bundle adjusted considering hybrid structure constraint. Zhang \etal~\cite{zhangsemantic} aligned the point clouds from vision and a semantic LiDAR map by a point-to-plane ICP algorithm utilizing semantic consistency.  Maplab 2.0 ~\cite{cramariuc2022maplab} extended the multi-session visual-inertial mapping framework \cite{schneider2018maplab} into a general multi-modal mapping system. Inter-session LiDAR registration can be used for loop-closure detection to align reference frames across sessions and mitigate drift.  Oishi \etal~\cite{oishi2020c,oishi2023lc} generated rendered image from the prior map, and the localization is performed in a simultaneous tracking and rendering manner.

Different from the above-mentioned methods, the proposed method doesn't require a known initial pose during localization, and LiDAR fusion is only performed at an offline mapping stage, keeping the online localization lightweight, easy to deploy but still accurate.  We outline the key challenges for this problem as follows:

\subsubsection{Cross-Modality} Establishing correspondence between LiDAR point cloud and images can be challenging due to the difference in representation, sensing range, and density.  
\subsubsection{Relocalization Ability}
Relocalization ability is key to a positioning system, as the initial pose for tracking in a prior map is hard to retrieve in practice, especially in a GPS-denied scenario. This ability is also essential for a positioning system to recover from localization failure~\cite{TK_loopclosures}.

\subsubsection{Map Representation} Operating on the raw point cloud is inefficient when searching for correspondences from the prior map. Map representation requires careful design based on the usage scenario and also the data association strategy. 
\subsubsection{Long-term Operation}  For long-term localization, the system may suffer from environment changes due to factors like lighting conditions, seasonal variations, construction etc., which may affect the data association reliability.
\subsubsection{Real-time Performance} Extra operations like depth estimation, data association, etc. limit the performance of pose estimation for large-scale outdoor applications, especially for embedded platforms with strict resource limits.

In this paper, we proposed a novel system using only a monocular camera to achieve comparable onboard localization performance by tracking deep-learning visual features on a LiDAR-enhanced visual prior map. Addressing the key challenges mentioned above, the proposed method has the following features:
\begin{itemize}
    \item Heavy computations are kept in an offline mapping stage and only lightweight tracking is performed during online localization to ensure real-time performance.
    \item  The keyframe poses with scale and a high-precision visual feature map are jointly recovered and optimized in a geometry-aware bundle adjustment problem, considering both visual and geometry constraints. 
    \item Volumetric representation enables efficient offline ray tracing for visual-LiDAR data association, while the co-visibility graph of keyframes facilitates quick online visual submap retrieval.
    \item A hierarchical localization system suitable for long-term, real-time applications, offering accurate and robust online re-localization and feature tracking.
\end{itemize}
    
The proposed localization system is not only evaluated on a benchmark dataset but is also deployed in a real-world autonomous driving platform, demonstrating the effectiveness of the proposed method.

%




\begin{figure}[t]
    \centering
    \vspace{3pt}
    \includegraphics[width=.38\textwidth]{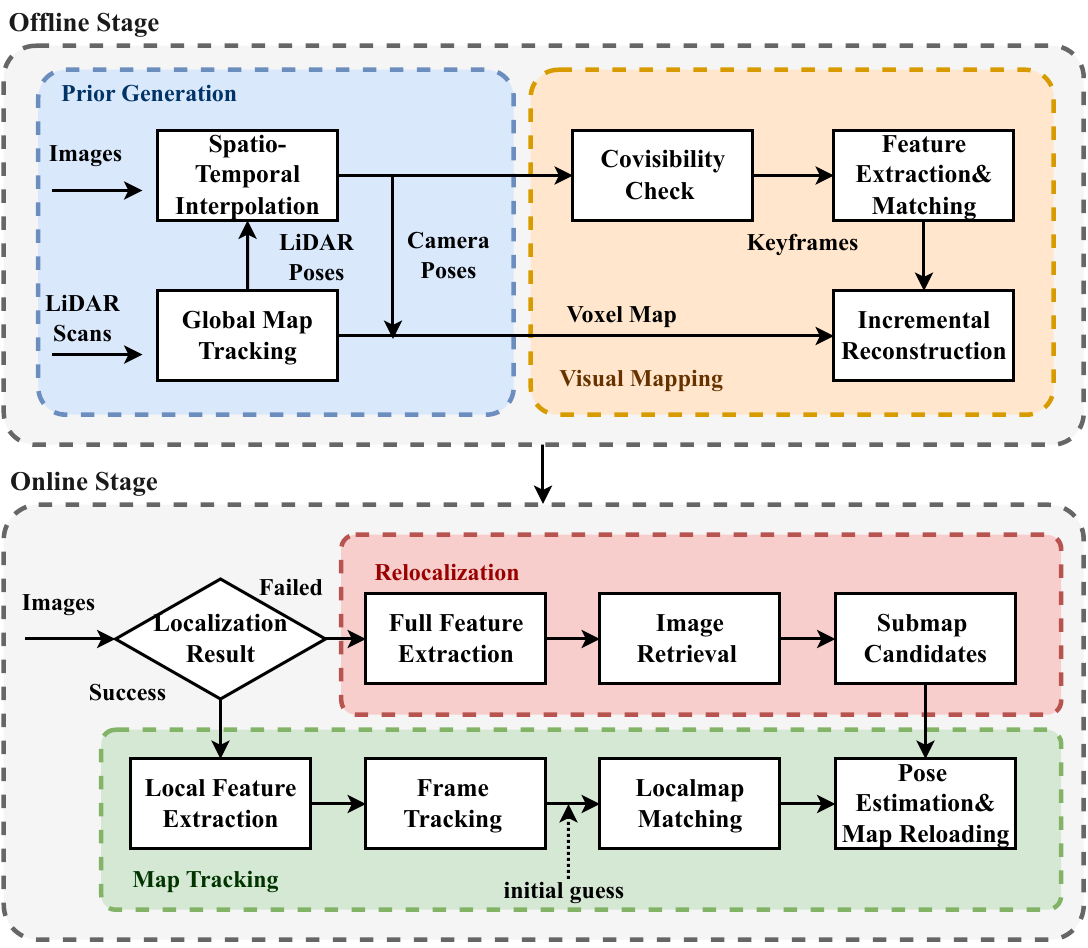}
    \setlength{\belowcaptionskip}{-15pt} 
    \caption{System pipeline of the proposed method.}
    \label{fig:pipeline_overview}
\end{figure}

\section{Proposed Method}

\subsection{Notation}

In the following part of this paper,
we use bold uppercase to denote matrix,  and bold lowercase for vector, e.g. $\mathbf{T} \in \mathrm{SE}(3)$ for rigid body transformation and $\mathbf{t} \in \mathbb{R}^3$ for translation. Rotation is defined by the rotation matrix
 $\mathbf{R} \in \mathrm{SO}(3)$ and quarternion $\mathbf{q}\in\mathbb{H}$.
 $\mathbf{p}= [x,y,z]^T \in \mathbb{R}^3$ denotes 3D point coordinate, $\mathbf{x}= [u,v]^T \in \mathbb{R}^2$ for 2D pixel coordinate. 
The coordinates and frame index are clarified at the superscript and subscript, e.g. the camera pose under the world frame at time $i$ is denoted as $\mathbf{T}^{\text{wc}}_i$. The brace $\{\cdot\}$ is introduced to represent a set of data, e.g. $ \{\mathbf{x}_j\} $ denotes a set of image pixels.
Other symbols and variables will be introduced and defined within their respective sections. 
\subsection{System Overview}

The architecture of the proposed method is illustrated in Fig. 
 \ref{fig:pipeline_overview}, where dashed rounded rectangle with different color represents different submodule of the system. The whole system includes an offline mapping stage and an online localization stage. During the offline mapping stage, our goal is to build a prior map, which is further used for localization purpose.
The offline stage is separated into two steps, the prior generation step which will be discussed in Sec.\ref{subsec:visprior}, and the visual mapping step which will be discussed in Sec.\ref{subsec:vismap}. During the online localization stage, the camera pose is estimated by efficiently tracking the image sequence on the prior map, which will be discussed in Sec.\ref{subsec:visloc}.

\subsection{Prior Generation}\label{subsec:visprior}

Prior generation takes LiDAR scans and monocular image sequence as input, and is responsible for providing pose priors of each image within the reference global point cloud map $\mathcal{M}^{\text{ref}}$. Meanwhile, it produces a voxel map of the operation area generated from the LiDAR scans, serving as geometry priors in the visual mapping stage. 

$\mathcal{M}^{\text{ref}}$ is first converted into K-D tree and point-wise covariance matrix $\mathcal{C}^{\text{ref}} = \{\mathbf{C}^{\text{ref}}_i \}$. Then for each incoming LiDAR scan $\mathcal{P}_k = \{\mathbf{p}_{k,i} \}$, we perform similar preprocessing step and get $\mathcal{C}^k$. 
 The transformation $\textbf{T}^{\text{wl}}_k$ of each LiDAR scan under global map frame can be solved by scan-to-scan (eq. \eqref{eq:s2s}) and scan-to-map (eq. \eqref{eq:s2m}) two-stage GICP (Generalized Iterative Closest Point) registration:
\vspace{-5pt}
\begin{equation}
\small
\begin{split}
    \hat{\textbf{T}}^{(k-1)}_{k}  =
    \argmin_{\textbf{T}^{(k-1)}_{k}}  \sum_i^N \mathbf{d}_i^\top \Big( 
     &\mathbf{C}^{(k-1)}_i   +  \\
     &\textbf{T}^{(k-1)}_k \ \mathbf{C}^k_i \  {\textbf{T}^{(k-1)}_k}^\top \Big) ^{-1} \mathbf{d}_i ,
    \label{eq:s2s}
\end{split}
\end{equation}
\vspace{-10pt}
\begin{equation}
\small
    \hat{\textbf{T}}_{k} = \argmin_{ \textbf{T}_{k}}  \sum_j^M \mathbf{d}_j^\top \left( \mathbf{C}^{\text{ref}}_j  +   \textbf{T}_{k} \ \mathbf{C}^k_j \  { \textbf{T}_{k}^\top} \right)^{-1} \mathbf{d}_j  .\\
    \label{eq:s2m}
\end{equation}
\noindent where $\mathbf{d}_i = \mathbf{p}_i^{k-1} - \textbf{T}^{(k-1)}_{k} \mathbf{p}_i^{k} $ and $\mathbf{d}_j = \mathbf{p}_j^{\text{ref}} - \textbf{T}_k \mathbf{p}_j^{k} $ 
indicate the point-wise distance error, and $\textbf{T}_k$ with $\textbf{T}_{(k-1)}\textbf{T}^{(k-1)}_k$ as initial guess. A detailed description of the GICP matching process can be referred to~\cite{chen2022dlo}. 

When the LiDAR sensor and camera is not hard synchronized, to get the 
prior pose of the mapping images, the interpolated pose  $ \textbf{T}_k = (\textbf{R}_k,\  \textbf{t}_k)$ at timestamp $k$ can be calculated by the nearest two neighbor frames at time $l$ and $r$ using the linear interpolation $\mathbf{t}_k = \mathbf{t}_l + s\cdot (\mathbf{t}_r - \mathbf{t}_l)$ for translation
 and spherical linear interpolation (slerp) $\textbf{R}_k = \text{slerp}(\textbf{R}_l, \textbf{R}_r, s)$ for rotation,
 with interpolation ratio $s= {(k - l)}/{(r - l)}$. The final interpolated LiDAR poses are further transformed to the camera frame using the sensor extrinsics $\mathbf{T}^\text{cl}$. 

Besides the camera trajectories, we also use the LiDAR sequence to build a reference UFOMap~\cite{ufomap} of the task area. This step is for two reasons, firstly rebuilding the reference map can make sure the prior data is up-to-date and consistent with the information from the mapping image sequence.  Also, this volumetric representation is more compact compared to raw point cloud and is superior in data loading and traversal. 

\subsection{Geometry-Aware Visual Mapping}\label{subsec:vismap}
In this step, the mapping images are fed into a SfM pipeline, along with the prior camera poses and the reference voxel map to recover the 3D geometry of the scene using visual features. 
\subsubsection{Incremental Reconstruction}
 We first sample keyframe images on the reference trajectory, ensuring a well-rounded understanding of the scene and also reducing data redundancy. If the relative translation and rotation between two keyframes fall below a specific threshold, they are considered to have co-observation, prompting visual matching for subsequent reconstruction. The SfM pipeline then leverages learning-based feature extractors SuperPoint~\cite{detone2018superpoint} to detect keypoints with local descriptors and NetVLAD~\cite{arandjelovic2016netvlad} to extract global image descriptors for similarity retrieval. Later, SuperGlue~\cite{sarlin2020superglue} is used for local feature matching of connected neighbors, enabling reconstruction based on the established visual feature correspondences. 

\begin{figure}[tb]
    \centering
    \vspace{5pt}
    \includegraphics[width=.3\textwidth]{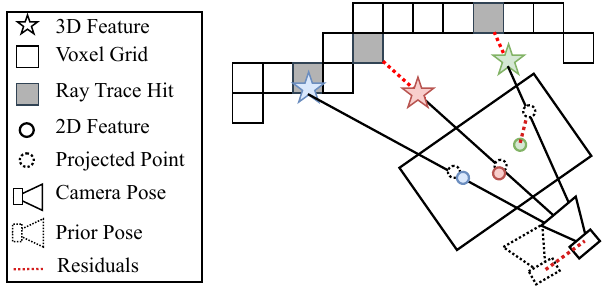}
    \setlength{\belowcaptionskip}{-16pt} 
    \caption{Illustration of the geometry-aware bundle adjustment.}
    \label{fig:vismapping}
\end{figure}

We then follow an incremental SfM scheme~\cite{schoenberger2016sfm,schoenberger2016mvs,wu2013linear} to reconstruct the visual map.
New images are progressively added to the problem by matching against the existing images followed by a BA step to refine the camera poses and 3D feature positions of an active image subset. The full reconstruction would be refined periodically by global optimization to further improve the reconstruction quality.

\subsubsection{Geometry-aware Bundle Adjustment (GABA)}
Since visual reconstruction using only monocular images has ambiguity in scale and may drift over time, in this paper, we propose to take full advantage of the prior information generated at Sec.\ref{subsec:visprior} by GABA to help recover scale and also improve the geometry consistency of the feature map against real-world structure. With $\mathcal{T}$  and $\mathcal{M^{\text{vis}}}$ be the set of camera poses and visual map point positions to be optimized, a GABA process can be modeled by an optimization problem:
\vspace{-5pt}
\begin{equation}
\label{eq:vismapping}
\small
\begin{split}
    \hat{\mathcal{T}}&,\mathcal{\hat{M}^{\text{vis}}}  = \underset{\mathbf{T}, \mathcal{M}_i}{\argmin} \sum_{i=1}^{\left\|\mathcal{T}\right\|} \sum_{j=1}^{\left\|\mathcal{M}_i\right\|} 
    \underbrace{ \rho \left( \left\| \mathbf{x}_j^i -\pi(\mathbf{T}_i, \mathbf{p}_j) \right\|_\Sigma \right)}_{visual\ factor} + \\
    &\underbrace{\rho \left( \left\| \mathbf{p}_j -\mathbf{r}(\bar{\mathcal{M}}^{\text{ref}},\mathbf{T}_i, \mathbf{p}_j) \right\|_\Sigma \right)}_{structure\ factor}+ \underbrace{\rho \left( \left\| \mathbf{e}(\mathbf{T}_i ,\mathbf{\Bar{T}}_i) \right\|_\Sigma \right)}_{prior\ factor}.
\end{split}
\end{equation}

\noindent where $\rho(\cdot)$ is the Huber norm to reduce the influence of outliers 
in the optimization, and $ \left\|x-\mu\right\|_\Sigma = (x-\mu)^T\mathbf{\Sigma}^{-1}(x-\mu)$ is the Mahalanobis norm to encounter the covariance of the residuals. For each image $I_i$, $\mathcal{M}_i$ denotes the set of visual map points visible at frame $i$.


The visual factor measures the disparity between observed position  $x_j^i$ of $\mathbf{p}_j$  at image $i$ and the corresponding projected position using projection function $\pi(\mathbf{T}_i, \mathbf{p}_j) = \mathbf{K}(\mathbf{T}_i^{-1} \cdot [\mathbf{p}_j^T,1]^T)$ with $\mathbf{K}$ denotes the intrinsic matrix of the camera.  
The structure factor measures the geometric difference between the estimated visual map $\mathcal{M^{\text{vis}}}$ and the prior LiDAR voxel map $\bar{\mathcal{M}}^{\text{ref}}$. 
The data association between the two maps can be found by an indirect or direct manner. In an indirect setting, map points are projected to the image plane according to the projection matrix. However, unlike LiDAR, cameras do not have a specified maximum sensing range, it is hard to cull the visible subarea of the prior map to process, making it inappropriate for large-scale outdoor scenes. In a direct setting, 
for each feature point $\mathbf{p}_j$, the corresponding reference structure is obtained by a ray tracing operation that simulates the path of a light ray starting from the camera center to the feature point and intersects with the reference map $\bar{\mathcal{M}}^{\text{ref}}$. The euclidean distance between the map point position $\mathbf{p}_j$ and the ray tracing hit point $\bar{\mathbf{p}_j} = \mathbf{r}(\bar{\mathcal{M}}^{\text{ref}},\mathbf{T}_i, \mathbf{p}_j)$  is added  as residual.
The ray tracing operation would be repeated after each optimization round to update data association. After several rounds of optimization, the reconstructed map would be geometrically consistent with the LiDAR map structure.

We also use the prior camera poses discussed in Sec.\ref{subsec:visprior} as anchor points to prevent 
the optimized pose going too far away from the initial guess. The prior pose residual can be modeled directly by the absolute difference between the estimated pose and the prior pose:
\begin{equation}
\small
\mathbf{e}(\mathbf{T}_i, \mathbf{\Bar{T}}_i)  = 
\begin{bmatrix}
    \Bar{\mathbf{t}} - \mathbf{t} \\
    2 \cdot \mathbf{q}^{-1} \Bar{\mathbf{q}}
\end{bmatrix} \in \mathbb{R}^6 .
\end{equation}

\begin{figure}[tb]
  \centering
  \vspace{5pt}

  \begin{subfigure}[t]{0.18\textwidth}
    \includegraphics[width=\textwidth]{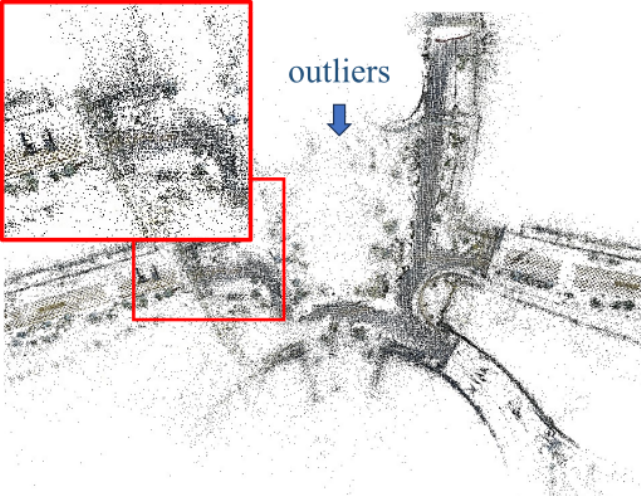}
    \captionsetup{skip=0.1pt}
    \caption{w/o. structure factor}
    \label{fig:renderb}
  \end{subfigure}
  \begin{subfigure}[t]{0.18\textwidth}
    \includegraphics[width=\textwidth]{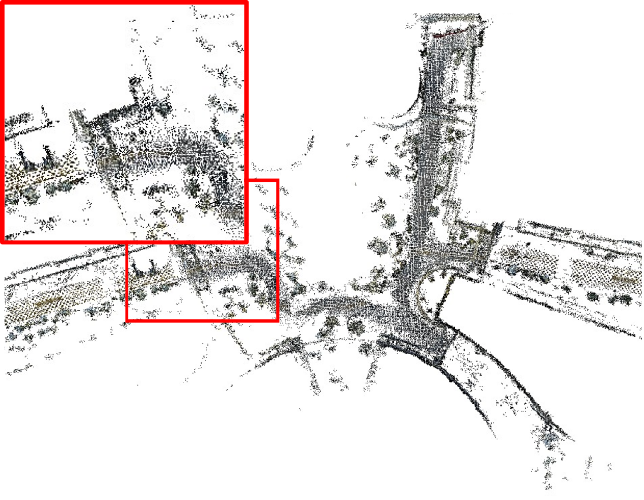}
    \captionsetup{skip=0.1pt}
    \caption{w/. structure factor}
    \label{fig:renderc}
  \end{subfigure}
  \captionsetup{skip=0.5pt}
  \setlength{\belowcaptionskip}{-15pt} 
  \caption{Reconstructed visual map w/o. and w/. structure factor. Introducing geometry constraints in visual mapping greatly improves the overall reconstruction quality.}
  \label{fig:render}
\end{figure}

Following the resolution of the GABA, the 3D positions of the inlier deep learning keypoints are subsequently refined. This optimization process ensures alignment of the reconstructed visual map $\mathcal{M^{\text{vis}}}$ with the global reference frame and also maintains both scale and geometric coherence with the prior map $\bar{\mathcal{M}}^{\text{ref}}$. Fig. \ref{fig:vismapping} gives an intuitive illustration of the GABA problem and a qualitative comparison of the visual mapping result is presented in Fig. \ref{fig:render}.

\subsection{Visual Localization} \label{subsec:visloc}
After the above visual mapping process, in this section, we will describe how to leverage the reconstructed visual prior map to get high-precision localization results.
As shown in Fig. \ref{fig:pipeline_overview}, during the online stage, the workflow goes into two branches, relocalization and map tracking, according to the current localization state.

The relocalization branch tries to provide pose estimation results when no initial guess is provided. Following the 
the hierarchical localization scheme~\cite{hloc2019}, the relocalization branch starts with extracting the global and local features $\mathcal{F}^g_i, \mathcal{F}^l_i = \{(\mathbf{x}^i_j, \mathbf{f}^i_j)\}$ of the input image $I_i$. The global features are used to retrieve $k$-nearest candidate keyframes and the corresponding connected clusters from the database. These candidate keyframes and connected cluster indicates the most possible positions of the input image. After that, the local features are used to perform keypoint matching between the input image and the candidate clusters. Since the 3D locations of the database keypoints are precomputed in the offline mapping stage, the final camera pose $\mathbf{T}_i$ can then be estimated using the PnP algorithm together with RANSAC outlier rejection. After the PnP-RANSAC optimization step, inlier matches are updated with the map point positions from the database and we get $\mathcal{F}^l_i = \{(\mathbf{x}^i_j, \mathbf{f}^i_j, \mathbf{p}_j)\}$.

 Successful relocalization leads to the camera tracking branch, where the input image's pose is recovered following a coarse-to-fine pose estimation scheme:
 \subsubsection{Frame Tracking}
\begin{figure}[bt]
  \centering
  \vspace{5pt}
  \begin{subfigure}[t]{0.18\textwidth}
    \centering
    \includegraphics[width=\textwidth]{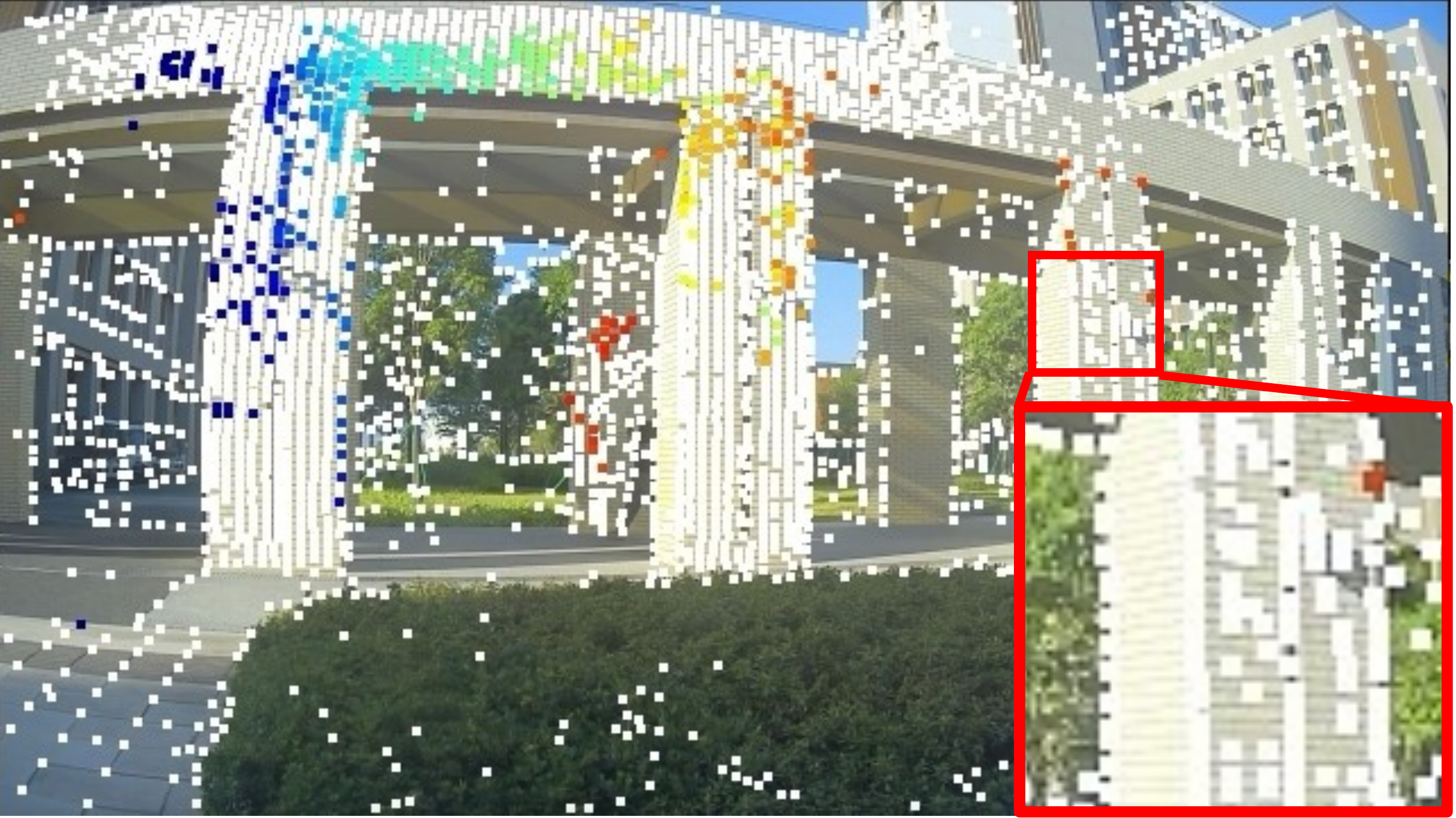}
    \captionsetup{skip=1pt}
    \caption{Tracking at time $t$}
    \label{fig:t}
  \end{subfigure}
  \hspace{-5pt}
  \begin{subfigure}[t]{0.18\textwidth}
    \centering
    \includegraphics[width=\textwidth]{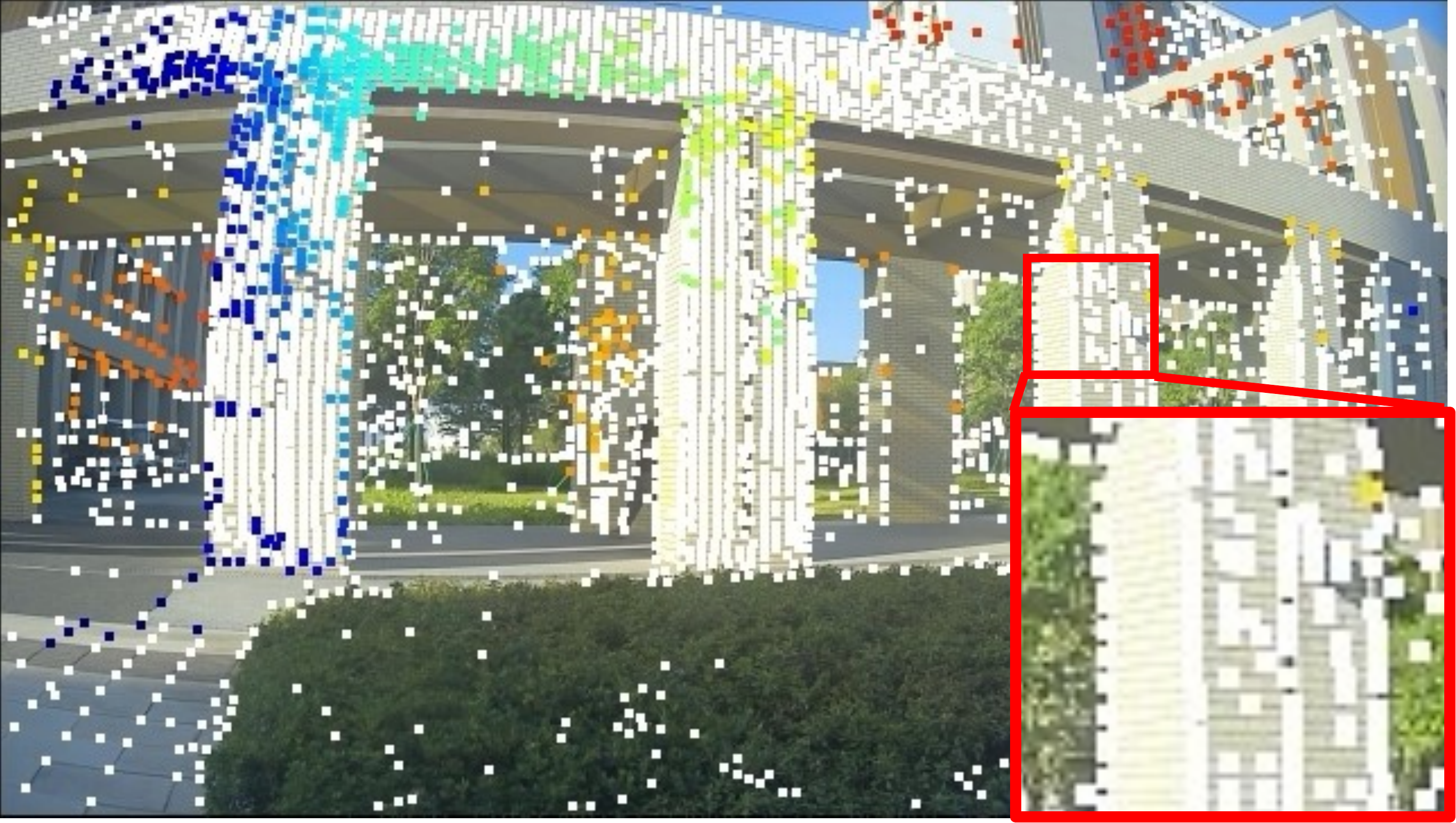}
    \captionsetup{skip=1pt}
    \caption{After map matching. }
    \label{fig:mapmatch}
  \end{subfigure}
  \begin{subfigure}[t]{0.18\textwidth}
    \centering
    \includegraphics[width=\textwidth]{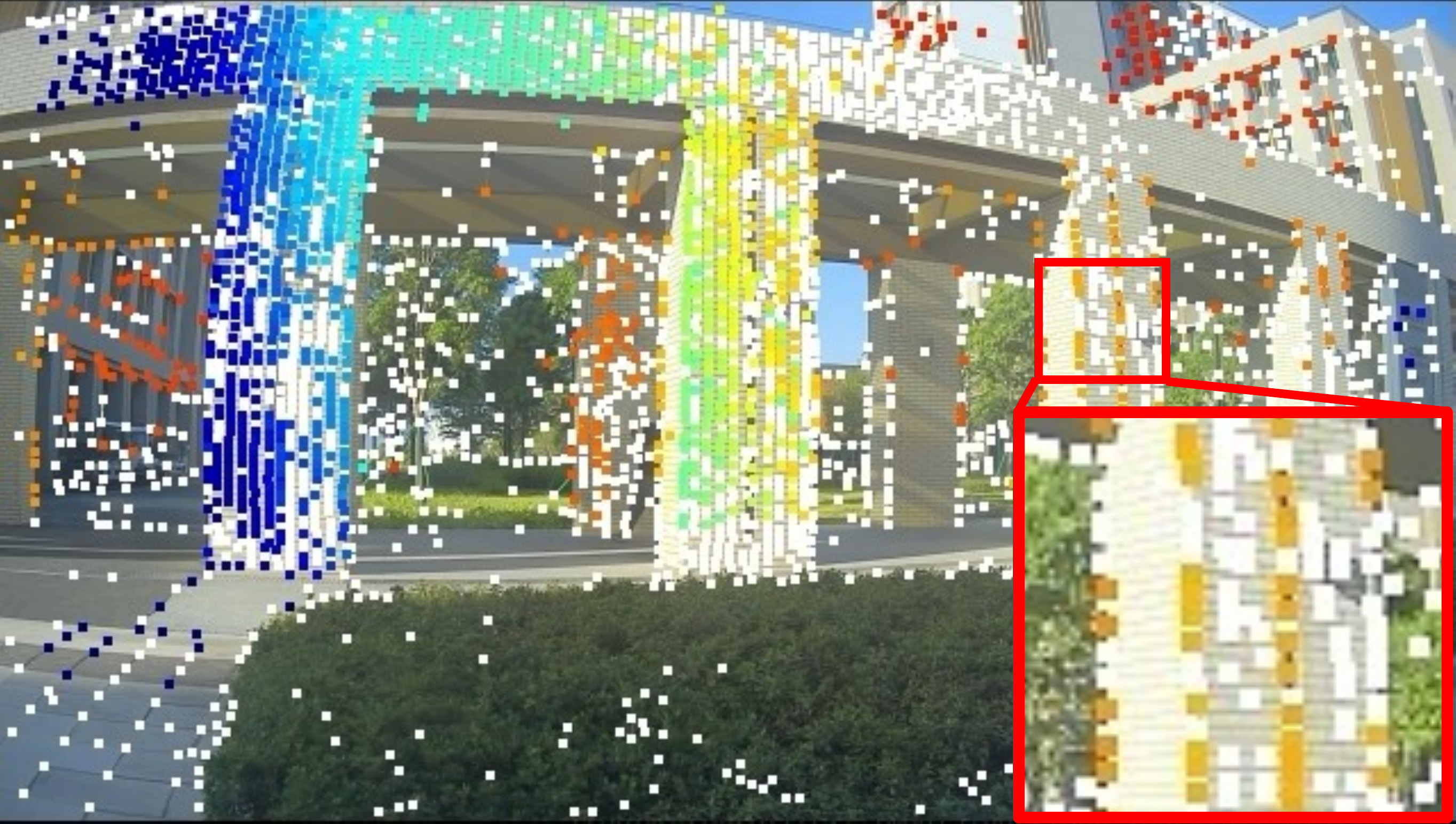}
    \captionsetup{skip=1pt}
    \caption{After map reload. }
    \label{fig:reload}
  \end{subfigure}
  \hspace{-5pt}
  \begin{subfigure}[t]{0.18\textwidth}
    \centering
    \includegraphics[width=\textwidth]{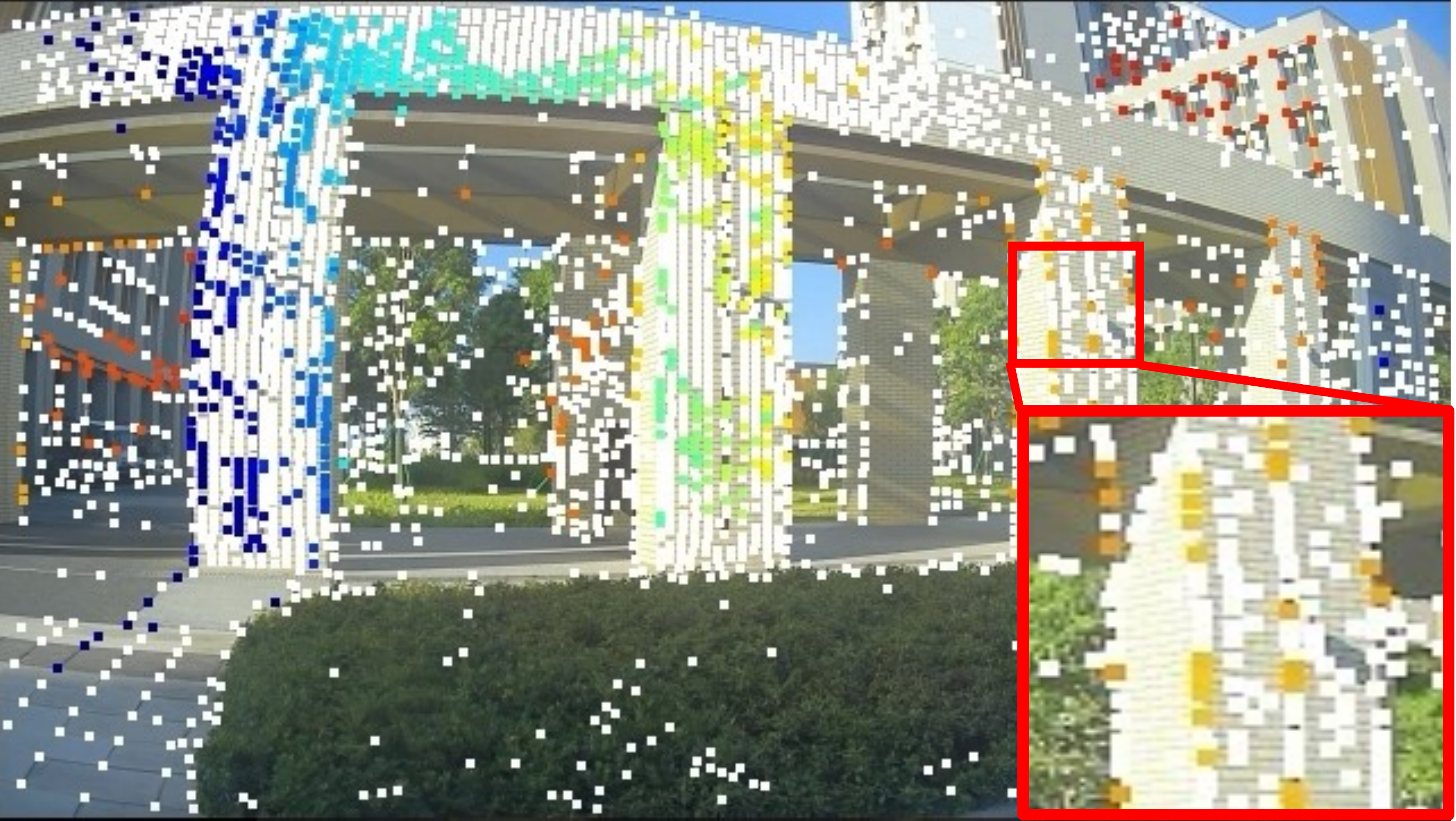}
    \captionsetup{skip=1pt}
    \caption{Tracking at time $t+1$}
    \label{fig:t1}
  \end{subfigure}
  \captionsetup{skip=2pt}
  \setlength{\belowcaptionskip}{-15pt} 
  \caption{Illustration of the feature tracking process. The tracked features are colored based on depth, free keypoints are colored by white. After the map reloading step, more map points that are not tracked at the map matching step are kept in the next frame.}
  \label{fig:tracking}
\end{figure}

 As shown in Fig. \ref{fig:tracking}, we first perform LK optical flow~\cite{lucas1981lk} estimation to track the pixel locations of local features $ \{(\mathbf{x}^{i-1}_j,  \mathbf{p}_j)\}$ from last frame into the current image plane $ \{(\bar{\mathbf{x}}^{i}_j, \mathbf{p}_j)\}$. Then we use these coarse map point associations to recover the camera pose $\bar{\mathbf{T}}_i$. This coarse estimation is then used as the prior pose of the following local map-matching step. 
 
\subsubsection{Local Map Matching}
The local map $\mathcal{M}^{i-1} = \{(\mathbf{f}_j, \mathbf{p}_j)\}$ consists of all the map points visible in the connected clusters $\mathcal{C}_{i-1}$ of last frame. We then need to establish data association between the latest local map  $\mathcal{M}^{i-1}$ and current frames' local features $\mathcal{F}^l_i = \{(\mathbf{x}^i_k, \mathbf{f}^i_k)\}$. Each associated map point-keypoint pair should satisfy two criteria: first, the keypoint pixel $\mathbf{x}^i_k$ should be near the reprojected pixel $\bar{\mathbf{x}}^i_j = \pi(\bar{\mathbf{T}}_i, \mathbf{p}_j)$, and second, the keypoint descriptor $\mathbf{f}^i_k$ should be similar to the map point descriptor $\mathbf{f}_j$, this can be modeled into a bi-parted graph matching problem. For each map point $\mathbf{p}_j$ we first use the reprojected pixel to search for features within a search radius $r$, and then associate it to the keypoint with the smallest descriptor difference.
\subsubsection{Pose Optimization and Map Point Reloading}
\begin{equation}
\label{eq:visloc}
\small
\begin{split}
    \hat{\mathbf{T}}_i  = \underset{\mathbf{\mathbf{T}_i}}{\argmin}  \sum_{j=1}^{\left\|\mathcal{M}^{i-1}\right\|} 
    &\underbrace{ \rho \left( \left\| \mathbf{x}_j^i -\pi(\mathbf{T}_i, \mathbf{p}_j) \right\|_\Sigma \right)}_{visual\ factor} +\\
    &\underbrace{\rho \left( \left\| \mathbf{e}(\mathbf{T}_i , \mathbf{\Bar{T}}_i) \right\|_\Sigma \right)}_{prior\ factor} 
\end{split} .
\end{equation} 
 After matching against the map frame, the camera pose of the current frame is refined by solving eq. \eqref{eq:visloc}, which is quite similar to eq. \eqref{eq:vismapping}, but only camera pose is optimized. The prior pose we use in eq. \eqref{eq:visloc} can directly come from the frame tracking step. 
 
 After the pose optimization, the outlier map point-keypoint matches would be removed ${\mathcal{F}^{-}}^l_{i} = \{(\mathbf{x}^i_k, \mathbf{f}^i_k, null)\}$, and for the inlier matches ${\mathcal{F}^{+}}^l_{i} = \{(\mathbf{x}^i_k, \mathbf{f}^i_k, \mathbf{p}_j)\}$, we update the map point descriptor in the database $\mathcal{M}^{i-1} = \{(\mathbf{f}_j, \mathbf{p}_j)\}$ by descriptor from current observation $\hat{\mathcal{M}}^{i-1} = \{(\mathbf{f}^i_k, \mathbf{p}_j)\}$ to ensure inter-frame continuity. Subsequently, as shown in Fig.~\ref{fig:tracking}, a refined local map matching operation with a very small search radius is performed to augment map point-keypoint candidates, enhancing localization by tracking more map points and ensuring system stability.

\section{EXPERIMENTAL RESULTS}
In this section, the proposed system will be assessed thoroughly in real-world captured and benchmark datasets to evaluate its performance and effectiveness in real-world applications.
\begin{figure}[tb]
 \centering
 \vspace{5pt}
 \begin{subfigure}[t]{0.19\textwidth}
    \centering
    \includegraphics[height=8.5em]{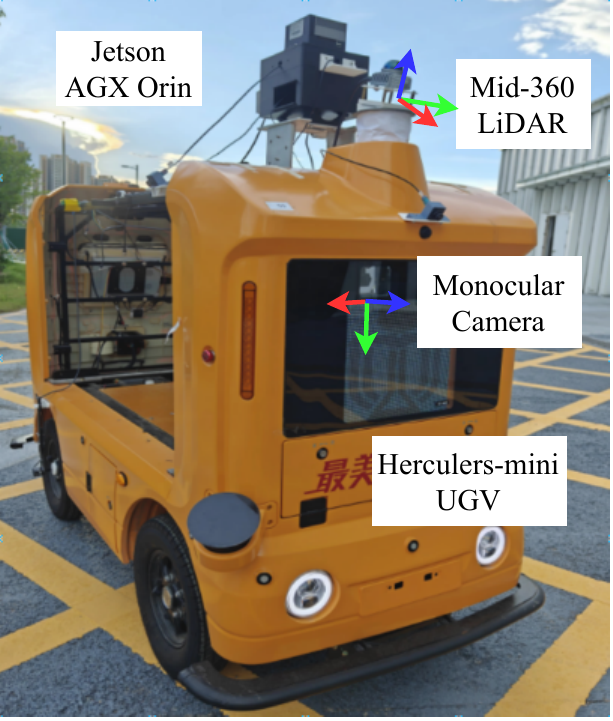}
    \caption{Sensor configuration. }
    \label{fig:sensor}
  \end{subfigure}
  \hspace{-22pt}
  \begin{subfigure}[t]{0.3\textwidth}
    \centering
    \includegraphics[height=8.5em, width=0.9\textwidth]{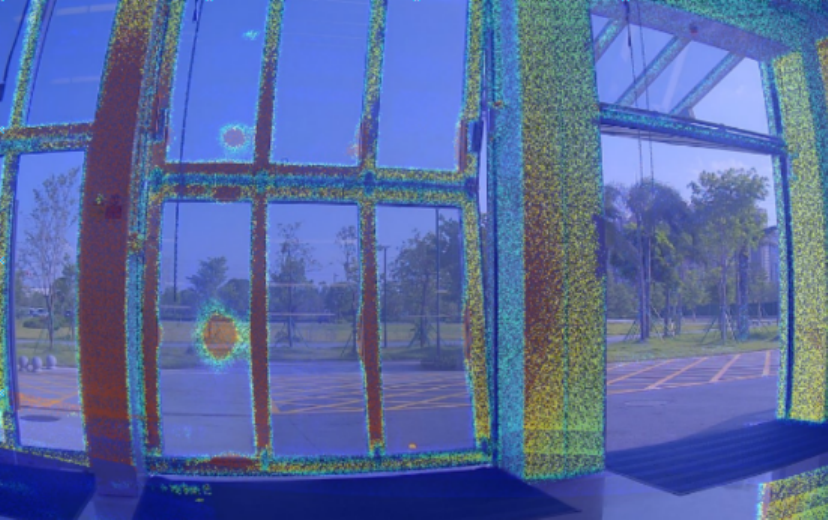}
    \caption{Calibration result by \cite{yuan2021pixel}.}
    \label{fig:platform}
\end{subfigure}
\setlength{\belowcaptionskip}{-18pt} 
\caption{Illustration of the experiment platform. The LiDAR is mounted $15^{\circ}$ front-facing to cover the camera's field of view.}
\label{fig:sensor-config} 
\end{figure}

\subsubsection{Campus Dataset}
As shown in Fig. \ref{fig:sensor-config}, a Hercules mini~\cite{hercules} carrying a Livox Mid-360 LiDAR and a monocular camera is used as the data acquisition platform. 
The LiDAR is ingeniously affixed to a detachable base, which can be unmounted from the system for mass and low-cost localization deployment once the prior map is built. 

The campus dataset contains $23$ sequences, each capturing $10$ Hz $1280\times720$ RGB images and $10$ Hz LiDAR scans. With a maximum driving speed of $15$ km/h and covering an approximate total mileage of $25$ km across $5$ distinct unmanned delivery operation routes, this dataset is specifically designed to facilitate navigation testing for robotic systems.

\subsubsection{KITTI Dataset}
To better evaluate the localization performance of our method, we use KITTI Odometry dataset~\cite{kitti1,kitti2} for benchmarking. The KITTI odometry dataset is captured by driving around the mid-size city of Karlsruhe, covering common outdoor automatic driving scenarios. The dataset contains time-synchronized undistorted stereo RGB images and LiDAR scans.

 The evo library~\cite{evo} is used for trajectory evaluation. Before each assessment, we perform $\mathrm{SE}(3)$ Umeyama alignment between the reference and estimated trajectories.

\subsection{Evaluation of All-day Localization Ability}
\begin{figure}[tb]
    \centering
    \vspace{5pt}
    \includegraphics[width=0.48\textwidth]{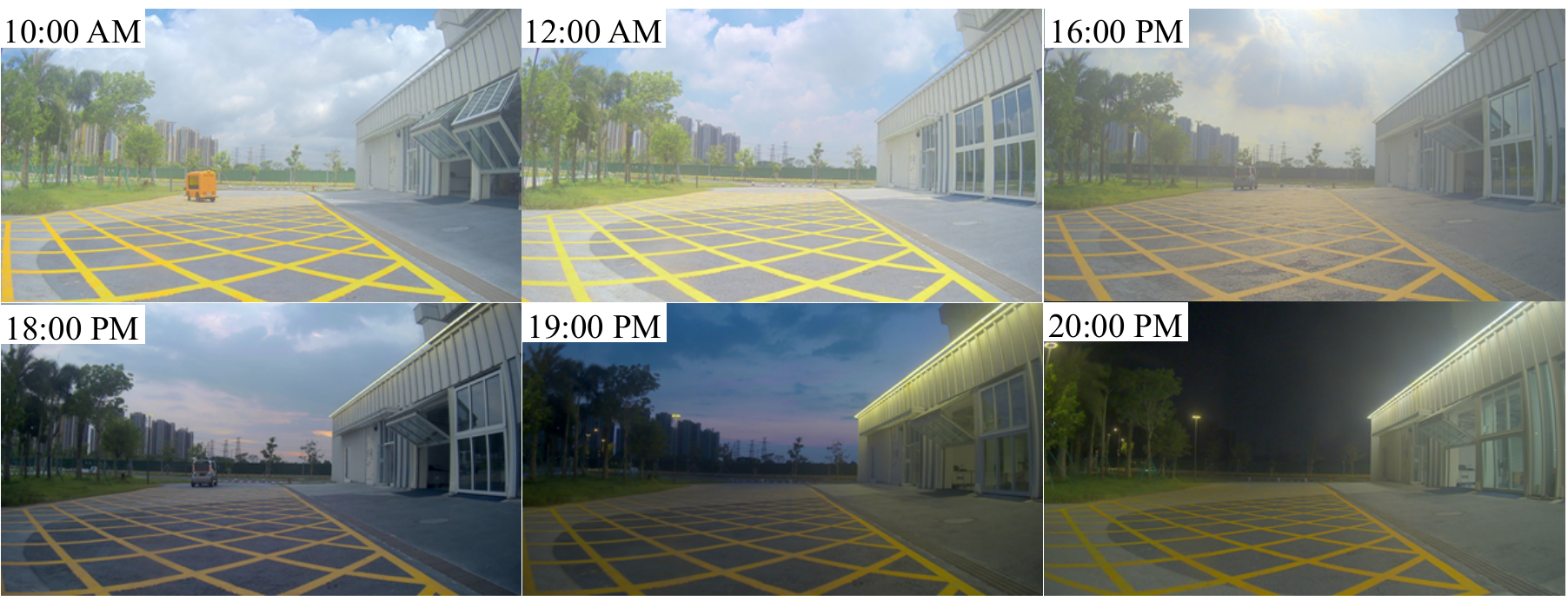}
    \caption{Illustration of the all-day localization dataset.}
    \label{fig:allday}
\end{figure}

Since visual features may behave inconsistently under different lighting conditions, we assess the all-day localization performance of the proposed method by evaluating its precision using prior maps captured at different hours within a day.
As illustrated in Fig. \ref{fig:allday}, several sequences on the same route at different moments are captured, remaining similar scene structures but showing different illumination. We went through different combinations of mapping and localization sequences to prove that the proposed method has a certain adaptability to scene illumination.

\begin{table}
\footnotesize
\centering
\setlength{\belowcaptionskip}{-8pt}
\caption{APE (m) for Different Combinations of Mapping and Localization Time (AM/PM)}
\label{tab:allday}
\sisetup{table-format=1.2e-1,table-number-alignment=center}
\begin{tabularx}{.458\textwidth}{
      c
      S[round-mode = places, round-precision=3, table-column-width=2em]
      S[round-mode = places, round-precision=3, table-column-width=2em]
      S[round-mode = places, round-precision=3, table-column-width=2em]
      S[round-mode = places, round-precision=3, table-column-width=2em]
      S[round-mode = places, round-precision=3, table-column-width=2em]
      S[round-mode = places, round-precision=3, table-column-width=2em]
      S[round-mode = places, round-precision=3, round-pad = false,table-column-width=2em]}
\toprule 
\backslashbox[5em]{\bf{$t_{\text{map}}$}}{\bf{$t_{\text{loc}}$}} & {\bf{10}}     & {\bf{12}}     & {\bf{15}}     & {\bf{16}}     & {\bf{18}}     & {\bf{19}}     & {\bf{20}}        \\ \midrule
\bf{10}                  & 0.0655 & 0.0639 & 0.0629 & 0.0667 & 0.0646 & 0.0685 & 1.6869     \\ 
\bf{12}                  & 0.0633 & 0.0542 & 0.0553 & 0.0569 & 0.0552 & 0.0589 & 0.0699     \\ 
\bf{15}                  & 0.0669 & 0.0567 & 0.0507 & 0.0536 & 0.0543 & 0.0571 & {-}  \\ 
\bf{16}                  & 0.0613 & 0.0547 & 0.0514 & 0.0526 & 0.0534 & 0.0579 & {-}   \\ 
\bf{18}                  & 0.0621 & 0.0569 & 0.0523 & 0.0523 & 0.0532 & 0.0517 & 4.0546     \\ 
\bf{19}                  & 0.0570 & 0.0543 & 0.0500 & 0.0525 & 0.0495 & 0.0477 & 0.0476     \\ 
\bf{20}                  & 4.1632 & 0.0564 & 0.0525 & 1.9331 & 0.3837 & 0.0501 & 0.0469     \\ \bottomrule
\end{tabularx}
\vspace{-20pt}
\end{table}
\begin{figure}[t]
  \centering
  \vspace{5pt}
  \begin{subfigure}[t]{0.18\textwidth}
    \includegraphics[width=\textwidth]{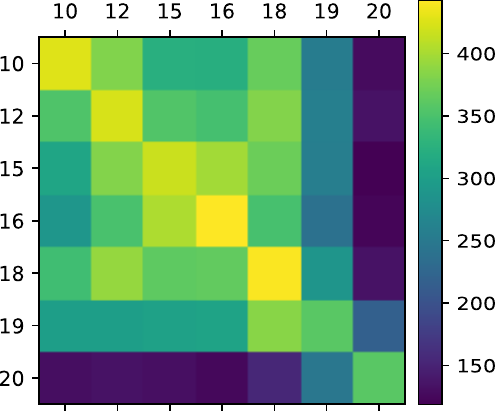}
    \caption{Average feature tracked.}
    \label{fig:alldc}
  \end{subfigure}
  \begin{subfigure}[t]{0.185\textwidth}
    \includegraphics[width=\textwidth]{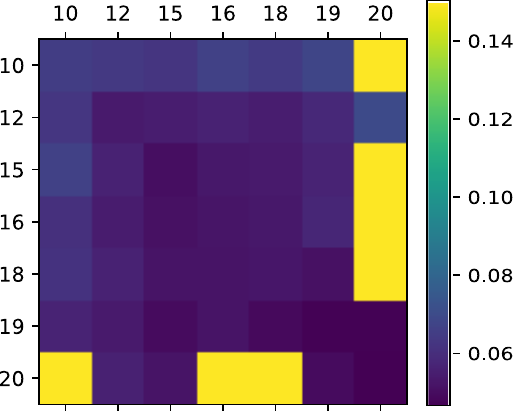}
    \caption{APE Mean (m).}
    \label{fig:allda}
  \end{subfigure}
  \setlength{\belowcaptionskip}{-5pt} 

  \caption{False color rendering of all-day localization results with mapping time on the vertical axis and localization time on the horizontal axis.}
  \label{fig:allmat}
\end{figure}

The mean absolute pose error (APE) results are presented in Table.\ref{tab:allday}. For better illustration, we organize the pairwise mean APE and also the average tracked features per frame in matrix form and visualize them by false color rendering (Fig.~\ref{fig:allmat}). The proposed method shows consistent localization ability on mapping sequences collected in most of the daytime unless it gets completely dark (around $20$ PM in our case, e.g. in test $20$--$10$ and test $15$--$20$ large pose error or tracking lost is observed, the average features tracked per frame is also low, possibly due to bad data associations).

\subsection{Evaluation on Campus dataset}

In this section, we evaluate the localization performance on four different scenes in our campus dataset.
The reference LiDAR map of the campus is pre-generated using a hand-held data acquisition platform~\cite{jiao2022fusionportable} and the localization accuracy is evaluated with respect to this global map. 

For each scene, we pick two mapping sequences and generate two visual maps. All sequences are used for localization on both maps if not chosen for mapping. LiDAR trajectories in the global map generated by Sec.\ref{subsec:visprior} serve as prior and ground truth for mapping and localization, respectively. The mean APE is calculated based on the experiments conducted for each scene.
We compare the proposed method with the hloc~\cite{hloc2019} toolkit, which contains similar global image retrieval and local feature matching settings but is designed for general visual localization purpose. The experiment result is presented in Table. \ref{tab:campus}. 
The original hloc pipeline failed to generate reasonable reconstruction results in scene $04$ and observed severe drift in scene $01$--$03$.  By combining hloc with the feature map generated by the proposed method (Sec. \ref{subsec:vismap}), the estimated absolute pose is reliable. Furthermore, the proposed hierarchical and lightweight localization method can achieve the same level of accuracy without using a learning-based feature matcher (e.g. Superglue\cite{sarlin2020superglue}, run only at around $2$ Hz for top-$10$ retrieved candidates of each query) and significantly reduce the mean relative pose error (RPE) at the same time.
This indicates our system is suitable for robotic navigation tasks, which require high localization accuracy while restricting computational resources.

\begin{table}
\footnotesize
\centering
\setlength{\belowcaptionskip}{-3pt}
\caption{Localization Comparison on Campus Dataset}
\label{tab:campus}
\sisetup{table-format=1.2,table-number-alignment=center}
\begin{tabularx}{.495\textwidth}{
      c
      S[round-mode = places,round-precision=3, table-column-width=3em]
      S[round-mode = places,round-precision=3, table-column-width=3em]
      S[round-mode = places,round-precision=3, table-column-width=3em]
      S[round-mode = places,round-precision=3, table-column-width=3em]
      S[round-mode = places,round-precision=3, table-column-width=3em]
      S[round-mode = places,round-precision=3, table-column-width=3em]}
\toprule
\multirow{2}{*}{\textbf{Scene}} & \multicolumn{2}{c}{{\textbf{hloc\cite{hloc2019}}}} & \multicolumn{2}{c}{\textbf{hloc + our map}} & \multicolumn{2}{c}{\textbf{Proposed}} \\ \cline{2-7} 
                                & {APE(m)}   & {RPE(m)}& {APE(m)}  & {RPE(m)} & {APE(m)}  & {RPE(m)} \\ \midrule
{01}      &97.983   & 1.9524    & \bfseries 0.1000 &  0.107           & 0.1133           & \bfseries 0.03479            \\ 
{02}      & 138.669 & 2.100     & 0.1605           & 0.1168           &\bfseries 0.1147  & \bfseries 0.05256    \\ 
{03}      & 75.198  & 1.7932    & \bfseries 0.0994 & 0.1007           & 0.1056           & \bfseries 0.04532    \\ 
{04}      &{-}      & {-}       & \bfseries 0.0953 & 0.0983           & 0.0994           & \bfseries 0.0364     \\ 
\bottomrule
\end{tabularx}
\vspace{-15pt}
\end{table}

\subsection{Evaluation on KITTI Dataset}
In this experiment, we first use the SemanticKITTI API~\cite{SemanticKITTI} to merge all the LiDAR scans into the reference point cloud map using loop-closured LiDAR trajectories. For workflow consistency, a similar LiDAR tracking procedure is done to generate the reference LiDAR trajectories for visual mapping purpose. Since the proposed method is mainly suitable for localization in repeated operation routes, the left color image would be used to generate the visual feature map and the right image would be used for localization test. Localization results would be compared with loop-closured LiDAR trajectories. 
We compare our method against \cite{kim2018stereo}, which also focuses on outdoor camera tracking using prior map and evaluated on the KITTI dataset except highway sequence $01$ due to the limited amount of features in the scene. The comparison result is presented in Table.\ref{tab:kitti}, and we reuse the result presented in \cite{kim2018stereo}. We can see that the proposed method can achieve centimeter-level accuracy on the KITTI dataset. It's worth pointing out that, our localization result is not compared against the provided ground-truth pose, because the prior point cloud map generated using ground-truth poses is of low quality. However, this experiment is sufficient to show that the localization result provided by our method has strong consistency against the prior map, which is more  
desirable in many autonomous driving applications as described in Sec. \ref{sec:intro}.
\begin{table}[tb]
\footnotesize
\vspace{5pt}
\caption{Localization Comparison on KITTI Dataset}\label{tab:kitti}
\centering
\setlength{\abovecaptionskip}{-3pt}
\sisetup{table-format=1.3,table-number-alignment=center}
\begin{tabularx}{.376\textwidth}{
      c
      S[round-mode = places, round-precision=3, table-column-width=3em]
      S[round-mode = places, round-precision=3, table-column-width=3em]
      S[round-mode = places, round-precision=3, table-column-width=3em]
      S[round-mode = places, round-precision=3, table-column-width=3em]
      }
\toprule
\multirow{2}{*}{\textbf{Sequence}} & \multicolumn{2}{c}{\textbf{Proposed}} & \multicolumn{2}{c}{{ \textbf{Kim~\etal~\cite{kim2018stereo}}}}          \\ \cline{2-5} 
                                   & {APE(m)}   & {APE($^\circ$)}& {APE(m)}  & {APE($^\circ$)}  \\ \hline
00                                 & \bfseries 0.048   & \bfseries 0.013   & 0.1325 & 0.3221 \\  
02                                 & \bfseries 0.149   & \bfseries 0.016   & 0.2205 & 0.3262 \\  
03                                 & \bfseries 0.031   & \bfseries 0.008   & 0.2368 & 0.4133 \\  
04                                 & \bfseries 0.041   & \bfseries 0.045   & 0.4496 & 0.8758 \\  
05                                 & \bfseries 0.035   & \bfseries 0.007   & 0.1462 & 0.3402 \\  
06                                 & \bfseries 0.042   & \bfseries 0.008   & 0.3753 & 0.8485 \\  
07                                 & \bfseries 0.040   & \bfseries 0.006   & 0.1305 & 0.4872 \\ 
08                                 & \bfseries 0.049   & \bfseries 0.007   & 0.1440 & 0.3279 \\  
09                                 & \bfseries 0.040   & \bfseries 0.007   & 0.1799 & 0.3375 \\  
10                                 & \bfseries 0.039   & \bfseries 0.006   & 0.2398 & 0.4934 \\ 
\bottomrule
\end{tabularx}
\end{table}

\subsection{Localization Runtime Analysis}
The localization runtime performance of the proposed system is measured in two different platforms: an  Intel Core i9-12900KF desktop with GeForce RTX 3080Ti and a Nvidia Jetson AGX Orin (32G). The feature extractors are speeded up using the Torch-TensorRT engine, and we use half-precision for Orin platform.  The time cost result is presented in  Table.\ref{tab:time}, in which the localization step is the sum of local feature extraction and tracking steps. The localization takes only $72$ ms on the Orin platform, which is sufficient for real-time localization of a 10 Hz image input and is suitable for onboard applications.

\begin{table}[tb]
\footnotesize
\centering
\setlength{\belowcaptionskip}{-5pt}
\setlength{\abovecaptionskip}{3pt}
\caption{Localization Time Cost (ms)}
\label{tab:time}
\begin{tabular}{l c c }
\toprule
\textbf{Steps}            & \textbf{Desktop} & \textbf{Orin} \\ \midrule 
Global Feature Extraction &  645             & 1180          \\ 
Relocalization            &  330             & 450           \\ \midrule
Local Feature Extraction  &  15              & 35            \\  
Tracking                  &  31              & 37            \\  
Localization              &  45              & 72            \\ \bottomrule
\end{tabular}
\vspace{-20pt}
\end{table}

\subsection{Unmanned Delivery Deployment}
The proposed localization system is adapted to an autonomous delivery system to track a preset delivery route. The proposed localization method can well cooperate with downstream modules like path planning and control, demonstrating that the proposed system can serve as a low-cost substitution for a LiDAR-based localization module. The result is presented in the attached media.
\section{Conclusions}
In this paper, a localization system using only a monocular camera is proposed. A high-quality visual prior map that is geometrically consistent with real-world structure is built by considering LiDAR structure constraint, prior pose constraint, and visual feature constraint in the proposed geometry-aware bundle adjustment problem. By tracking deep-learning features on this prior map, high-precision localization results are provided with a real-time guarantee, which is favorable for low-cost robot system deployment in real-world applications.  

In future work, the proposed method will be further optimized by integrating IMU in a tightly coupled configuration and also adapting the current pipeline to multi-camera systems or panoramic cameras to improve the robustness in challenging featureless scenarios.

\clearpage
\bibliographystyle{unsrt}
\bibliography{ref}

\end{document}